\documentclass[11pt]{article}

\usepackage[margin=1in]{geometry}
\usepackage{amsmath,amsfonts,amsthm,amssymb}
\usepackage{graphicx}
\usepackage{booktabs}
\usepackage{xcolor}
\usepackage{threeparttable}
\usepackage[round]{natbib}
\usepackage{caption}
\usepackage{url}
\usepackage{algorithm}
\usepackage{algorithmic}
\usepackage{newfloat}
\usepackage{listings}

\newtheorem{proposition}{Proposition}
\newtheorem{remark}{Remark}

\theoremstyle{plain}
\makeatletter
\newtheorem*{rep@theorem}{\rep@title}
\newcommand{\newreptheorem}[2]{%
  \newenvironment{rep#1}[1]{%
     \def\rep@title{#2 \ref{##1}}%
     \begin{rep@theorem}
  }%
  {\end{rep@theorem}}%
}
\makeatother

\newreptheorem{theorem}{Theorem}
\newreptheorem{lemma}{Lemma}
\newreptheorem{proposition}{Proposition}
\newreptheorem{corollary}{Corollary}

\DeclareCaptionStyle{ruled}{labelfont=normalfont,labelsep=colon,strut=off}
\floatstyle{ruled}
\newfloat{listing}{tb}{lst}
\floatname{listing}{Listing}

\newenvironment{links}{\bigskip\noindent}{\par\bigskip}
\newcommand{\link}[2]{\textbf{#1: } \url{#2}}

\title{Reward Redistribution via Gaussian Process Likelihood Estimation}

\author{Minheng Xiao\thanks{Department of Integrated Systems Engineering, The Ohio State University, Columbus, OH, USA, Email: {\tt xiao.1120@osu.edu};}~~~
Xian Yu\thanks{Corresponding author; Department of Integrated Systems Engineering, The Ohio State University, Columbus, OH, USA, Email: {\tt yu.3610@osu.edu}}}

\date{}

\begin{document}

\maketitle

\begin{abstract}
In many practical reinforcement learning tasks, feedback is only provided at the end of a long horizon, leading to sparse and delayed rewards. Existing reward redistribution methods typically assume that per-step rewards are independent, thus overlooking interdependencies among state–action pairs. In this paper, we propose a Gaussian Process-based Likelihood Reward Redistribution (GP-LRR) framework that addresses this issue by modeling the reward function as a sample from a Gaussian Process (GP), which explicitly captures dependencies between state–action pairs through the kernel function. By maximizing the likelihood of the observed episodic return via a \textit{leave-one-out} strategy that leverages the entire trajectory, our framework inherently introduces uncertainty regularization. Moreover, we show that the conventional mean squared error (MSE)-based reward redistribution arises as a special case of our GP-LRR framework when using a degenerate kernel without observation noise. When integrated with an off-policy algorithm such as Soft Actor-Critic, GP-LRR yields dense and informative reward signals, resulting in superior sample efficiency and policy performance on several MuJoCo benchmarks.
\end{abstract}

\begin{links}
    \link{Code}{https://github.com/xiao-1120/AAAI-LRR}
\end{links}

\section{Introduction}
Reinforcement learning (RL) has been successfully applied in a wide range of fields, including finance \cite{hambly2023recent}, healthcare \cite{yu2021reinforcement}, robotics \cite{kober2013reinforcement},
and autonomous driving \cite{kiran2021deep}. However, a critical challenge remains: in many practical systems, feedback is provided only at the end of an episode after a long sequence of actions, rather than immediately. For example, in aerospace design, the quality of an aircraft component is evaluated only after the complete manufacturing process is finished \cite{razzaghi2024survey}. This delayed feedback creates an extremely sparse reward landscape, making it difficult to determine which actions most significantly influenced the final outcome. Consequently, the learning process may converge slowly or become trapped in suboptimal policies. Overcoming this bottleneck is essential for developing more robust and efficient RL solutions across diverse application domains.

\begin{figure}
\centering
\begin{tabular}{cc}
\hspace{-0.35cm}\includegraphics[width=0.22\textwidth, height=0.16\textwidth]{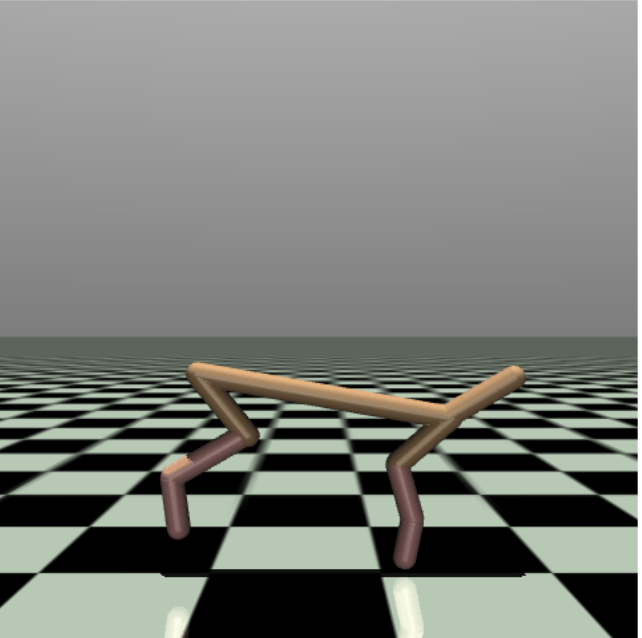} & 
\raisebox{-0.25cm}{\includegraphics[width=0.22\textwidth, height=0.18\textwidth]{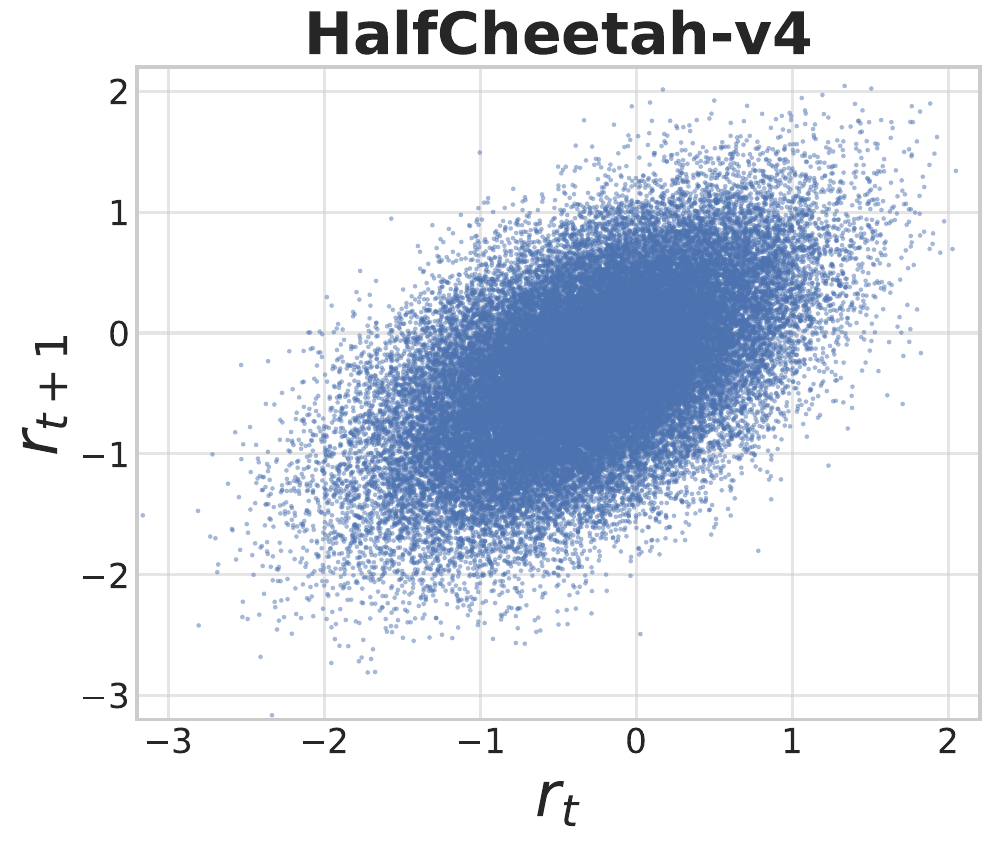}}
\end{tabular}
\caption{(Left) HalfCheetah-v4 environment visualization. (Right) Lag-1 autocorrelation in rewards collected under random policy, where each point represents a $(r_t, r_{t+1})$ pair from trajectory rollouts.}
\label{fig:example}
\end{figure}

Recent studies have started to address these issues from different perspectives~\citep{arjona2019rudder, gangwani2020learning, ren2021learning}. A widely adopted strategy to tackle sparse rewards is to decompose the episodic return into per-step contributions (credits), thereby generating dense reward signals. For example, RUDDER uses an LSTM to predict per-step returns and redistributes credit through backpropagation. The model is trained by minimizing the mean squared error (MSE) between predicted and actual returns.
RRD~\citep{ren2021learning} samples random trajectory subsequences to approximate optimal decomposition, and IRCR~\citep{gangwani2020learning} simply distributes returns uniformly across all timesteps. However, conventional return decomposition approaches typically assume that each step's reward is independent from each other. This assumption overlooks the interdependencies between state-action pairs on a trajectory in many tasks. A simple empirical evidence shows significant lag-1 autocorrelation in several environments (see Figure~\ref{fig:example}), indicating that rewards are interdependent over time due to the correlation among state-action pairs in consecutive steps. Ignoring such dependencies can lead to ineffective reward redistribution, where important interactions between actions are missed, eventually impairing learning efficiency.

To model such interdependence in the per-step rewards between different state-action pairs, Gaussian Process (GP) is a useful tool. Unlike the aforementioned regression approaches that assume the reward function to lie in a parametric family, GP is a nonparametric approach that treats the unknown reward function as a black box and defines a distribution over it. Through the kernel function (covariance), GP explicitly models the correlations of reward functions between different state-action pairs. In prior works, GP has most commonly been employed either for value-function approximation 
\citep{engel2005reinforcement, chung2013gaussian} or to model the transition dynamics \citep{deisenroth2011pilco}. 
Recently, \cite{wei2024safe} uses GP to approximate the cost function and provides regret bounds on safe RL with instantaneous constraints. However, to the best of our knowledge, this is the first work that leverages GP for reward redistribution in RL.

In this paper, we propose a novel GP-based reward redistribution framework. Instead of treating per-step rewards as isolated signals, our method assumes that rewards arise from a latent function with structured correlations captured by a kernel function. By leveraging a \textit{leave-one-out} (LOO) strategy, our framework computes the likelihood of the observed total return while naturally incorporating dependencies between state-action pairs across the entire trajectory. Our main contributions can be summarized as follows.
\begin{itemize}
    \item We introduce a GP-based Likelihood Reward Redistribution framework (GP-LRR) that models per-step reward as a sample from a GP and maximizes the likelihood of a trajectory based on LOO targets. This framework explicitly models the interdependent structure of per-step rewards and does not assume any parametric family for the reward function.
    \item We provide theoretical analyses showing that (i) GP-LRR generalizes existing MSE-based approaches, (ii) through the kernel function and precision matrix, GP-LRR explicitly captures the correlations between different state-action pairs and pools their prediction errors together to update the parameters, and (iii) GP-LRR implicitly regularizes reward estimates via its predictive uncertainty, avoiding overconfident assignments in poorly explored regions.
    \item We develop a practical algorithm that integrates GP-LRR with Soft Actor-Critic (SAC), demonstrating superior sample efficiency and asymptotic performance compared to state-of-the-art baselines on several MuJoCo benchmarks.
\end{itemize}

\subsection{Related Works}
We review the most relevant approaches for handling sparse and delayed rewards in RL through reward redistribution.

\textbf{LIRPG}~\citep{zheng2018learning} addresses sparse rewards by jointly learning a policy and an intrinsic reward. Meta-gradients ensure the intrinsic signal improves the true objective, but the on-policy requirement limits sample efficiency.

\textbf{GASIL}~\citep{guo2018generative} uses a GAN-style discriminator trained on high-return trajectories to provide dense rewards. It yields continuous learning signals but is on-policy and prone to mode collapse, hindering broader exploration.

\textbf{RUDDER}~\citep{arjona2019rudder} converts episodic returns into dense per-step signals via LSTM-based return decomposition, minimizing MSE to the true return. Long-horizon BPTT is unstable and costly, and complexity grows with trajectory length.

\textbf{IRCR}~\citep{gangwani2020learning} uniformly distributes the episodic return across timesteps. It is computation-light but ignores temporal structure, failing to isolate the truly responsible actions.

\textbf{RRD}~\citep{ren2021learning} samples short trajectory subsequences and minimizes a variance-regularized objective, balancing cost and accuracy. It integrates well with off-policy RL, but still ignores state–action dependencies and lacks a global view.

\textbf{GRD}~\citep{zhang2023interpretable} learns a causal generative model over states, actions, and latent Markovian rewards, proving identifiability. It produces interpretable, policy-invariant proxy rewards and compact state representations, outperforming IRCR/RRD on delayed-reward MuJoCo with SAC.

\textbf{DIASter}~\citep{lin2024episodic} decomposes return by cutting a trajectory into two sub-trajectories and defining step-wise rewards as expected differences of assigned sub-trajectory returns. It offers return-equivalence and near-optimal guidance guarantees, uses a GRU implementation~\citep{cho2014learning}, and improves sample efficiency on MuJoCo.

Overall, the above methods either rely on on-policy learning with limited sample efficiency, ignore interdependencies among state–action pairs during return redistribution, or employ heavy sequential neural networks over $(s,a)$ that incur substantial computational burden.

\section{Preliminaries}
\subsection{MSE-based Reward Redistribution}
Traditional RL approaches typically assume that the environment can be modeled as a Markov decision process (MDP) characterized by the tuple $\mathcal{M} = \langle \mathcal{S}, \mathcal{A}, P, R, \mu \rangle$, where \(\mathcal{S}\) and \(\mathcal{A}\) denote the state and action spaces, \(P(s' | s,a)\) captures the unknown transition dynamics, and \(R(s,a)\) is the reward function. The initial state distribution is given by \(\mu\). The goal is to learn a policy that maximizes the expected discounted sum of rewards:
\begin{align}
\label{eq:RL-objective}
J(\pi) = \mathbb{E}^{\pi}_{s\sim\mu}\left[\sum_{t=0}^{\infty} \gamma^t\, R(s_t, a_t) \mid s_0=s\right],
\end{align}
where $s_t\sim P(\cdot|s_{t-1},a_{t-1}),\ a_t\sim\pi(\cdot|s_t)$ and $\gamma\in(0,1)$ is the discount factor. However, in many real-world tasks, the per-step reward signal $R(s,a)$ remains unknown and the feedback is only provided at the end of an episode. Let \(\tau = \{s_0, a_0, s_1, \dots, s_T\}\) be a trajectory with a cumulative (or episodic) reward \(R_{\text{ep}}(\tau)\). In this episodic setting, the objective becomes
\begin{align*}
J_{\text{ep}}(\pi) = \mathbb{E}\left[R_{\text{ep}}(\tau) \,\Big|\, \tau \sim \pi, P\right].
\end{align*}
Because intermediate rewards are not provided, the decision process loses its strict Markovian property. Nonetheless, it is often assumed that the overall return can be approximately decomposed additively as
\begin{align*}
R_{\text{ep}}(\tau) \approx \sum_{t=0}^{T-1} R_b(s_t,a_t),
\end{align*}
where $R_b(s,a)$ is the underlying (unknown) per-step reward function.

To overcome the challenges associated with sparse episodic feedback, reward redistribution techniques aim to derive a surrogate reward function for $R_b(s,a)$ that transforms the delayed signal into a dense, step-by-step reward. A popular method to achieve this is through \emph{return decomposition} \cite{arjona2019rudder,efroni2021return}. In this approach, one trains a parameterized reward model \(R_{\theta}(s,a)\) by minimizing the mean squared loss 
\begin{align}
\label{eq:MSERewardRedistribution}
L_{\text{RD}}(\theta) = \mathbb{E}_{\tau \sim D} \left[ \left( R_{\text{ep}}(\tau) - \sum_{t=0}^{T-1} R_{\theta}(s_t,a_t) \right)^2 \right],
\end{align}
where $D$ is the dataset of collected trajectories. By minimizing this loss, the model is encouraged to assign local rewards that, when summed over the trajectory, closely match the true episodic return. Once a reliable proxy $R_{\theta}(s,a)$ is obtained, it can be used in the downstream policy optimization task to replace the sparse terminal reward with a more informative, dense signal, thereby accelerating policy optimization in environments with delayed rewards.

However, such approaches treat the per-step rewards as independent signals and impose strong assumptions on the parametric family. In this paper, we propose a non-parametric approach, GP-LRR, that explicitly models the correlations between different rewards and updates the parameters by maximizing the likelihood. We will demonstrate that GP-LRR encompasses the MSE approaches as special cases.

\subsection{Soft Actor-Critic}
Soft Actor–Critic (SAC)~\citep{haarnoja2018soft} is an off-policy Actor–Critic algorithm that augments the standard RL objective with an entropy term. Instead of optimizing the objective~\eqref{eq:RL-objective}, SAC seeks a policy $\pi$ that maximizes
\begin{align*}
J_{\text{SAC}}(\pi) = \mathbb{E}_{s\sim\mu}^{\pi}\bigg[\sum_{t=0}^{\infty}\gamma^t(R(s_t, a_t) + \alpha\mathcal{H}(\pi(\cdot|s_t)))\mid s_0=s\bigg],
\end{align*}
where $\mathcal{H}(\pi(\cdot|s_t)) = -\mathbb{E}_{a\sim\pi(\cdot|s_t)}[\log\pi(a|s_t)]$ is the Shannon entropy of the policy at state $s_t$ and the temperature $\alpha > 0$ controls the reward-versus-exploration trade-off. In the policy-evaluation stage, it updates a soft Q-function $Q_\omega$ by minimizing the mean-squared soft Bellman residual
\begin{align}
\label{eq:sac_critic}
\mathcal{L}_Q(\omega) = \mathbb{E}_{\substack{(s,a,r,s')\sim\mathcal D\\a'\sim\pi_\zeta}} \biggl[\frac{1}{2}\Bigl(Q_\omega(s, a) - \hat{Q}_\omega(s, a)\Bigr)^{2}\biggr],
\end{align}
where $\hat{Q}_{\omega}(s, a)$ is the soft TD-target defined as
\begin{align}
\label{eq:SAC-target}
\hat{Q}_{\omega}(s, a) = r + \gamma(Q_{\bar{\omega}}(s', a') - \alpha\log\pi_{\zeta}(a'|s')),
\end{align}
where $Q_{\bar{\omega}}$ is a slowly polyak-averaged target network and $\mathcal{D}$ is the replay buffer. In the policy-improvement stage, SAC performs a gradient step which minimizes the objective
\begin{align}
\label{eq:sac_actor}
{J_{\text{actor}}}(\zeta) = \mathbb{E}_{s\sim\mathcal{D}, a \sim \pi_{\zeta}(\cdot|s)}\big[\alpha\log\pi_{\zeta}(a|s) - Q_{\omega}(s, a)\big].
\end{align}
Since the optimal temperature depends on how stochastic one wishes the final policy to be, SAC learns the temperature $\alpha$ online by minimizing
\begin{align}
\label{eq:sac_temp}
{J_{\text{temp}}}(\alpha) = \mathbb{E}_{s \sim \mathcal{D}, a \sim \pi_{\zeta}(\cdot|s)}\big[-\alpha\log\pi_{\zeta}(a|s) - \alpha\bar{\mathcal{H}}\big],
\end{align}
so that the realized entropy matches a user-specified target $\bar{\mathcal{H}}$. By embedding the entropy bonus directly in both the critic target and the actor loss, SAC achieves high sample-efficiency, remains robust to function-approximation error, and attains state-of-the-art performance on continuous-control benchmarks.

We will integrate GP-LRR with SAC by using the current estimated mean function as the reward signal for SAC updates. While collecting trajectories with SAC, we periodically train the GP model on complete episodes to learn the reward redistribution and then use these redistributed rewards instead of the sparse episodic signal in SAC's Q-function and policy updates.

\section{GP-based Likelihood Reward Redistribution}
In this section, we present a GP-based reward redistribution framework that leverages probabilistic modeling and nonparametric regression. Unlike conventional methods that rely on parametric regression approaches~\cite{arjona2019rudder} or subsample segments of trajectories~\cite{ren2021learning}—both of which treat rewards as isolated, independent signals, our method models the per-step proxy reward as a random variable from a GP and computes the likelihood over the entire trajectory using the LOO strategy. We then update the parameters in the GP by maximizing the likelihood. As we will show in the Theoretical Analyses section, our framework includes the conventional MSE-based reward redistribution~\eqref{eq:MSERewardRedistribution} as a special case and inherently incorporates an uncertainty regularizer that robustly mitigates trivial solutions. Furthermore, GP-LRR enables gradient information pooling through explicit correlation modeling.

\subsection{Gaussian Process Likelihood}
In this section, we present a principled reward redistribution framework based on GP likelihood estimation. Our method explicitly models correlations between different state-action pairs through a kernel function, enabling effective credit assignment across temporally and spatially related states. The key insight is that similar state-action pairs should yield similar rewards—a structure we exploit through the GP's kernel function. 

\subsubsection{Gaussian Process Reward Modeling}
\label{subsec:GPRewardModeling}

Consider the per-step reward $R(s,a)$ at each state-action pair as a random variable following
\begin{align}
R(s, a) = R_b(s, a) + \epsilon, \quad \epsilon \sim \mathcal{N}(0, \sigma_\epsilon^2),
\end{align}
where $R_b(s, a)$ is the unknown ground-truth reward function at $(s,a)$ and $\epsilon$ represents the Gaussian noise. Instead of assuming the ground truth reward model $R_b(s, a)$ to lie in a parametric family, we model this as a black-box function and a possible sample from a GP:
\begin{align*}
R_b(s, a) \sim \mathcal{GP}\big(\mu_{\boldsymbol\theta}(s, a), k_{\boldsymbol\phi}((s, a), (s', a'))\big),
\end{align*}
where $\mu_{\boldsymbol\theta}(s, a)$ is the mean function of the reward at state-action pair $(s,a)$ parameterized by $\boldsymbol{\theta}$, and $k_{\boldsymbol\phi}((s, a), (s', a'))$ is the kernel function (covariance) between any two state-action pairs $(s, a)$ and $(s', a')$ with hyperparameters $\boldsymbol{\phi}$. Note that in the case of discrete state and action spaces, the mean function can be exactly represented by a lookup table $\mu(s,a)$ without any function approximation. To accommodate large-scale state/action space and continuous control environments, we will continue with the parameterized function $\mu_{\boldsymbol\theta}(s, a)$ and utilize neural networks to represent it in our experiments. To measure the correlation between different state-action pairs, one option is to use the Radial Basis Function (RBF) kernel~\citep{duvenaud2014kernel}:
\begin{align}
\label{eq:RBFKernel}
k_{\boldsymbol\phi}((s,a), (s',a')) = \sigma_f^2 \exp\left(-\frac{\|(s,a) - (s',a')\|^2}{2\ell_{\text{rbf}}^2}\right),
\end{align}
where $\boldsymbol\phi = [\sigma_f^2, \ell_{\text{rbf}}]$ with $\sigma_f^2$ being the signal variance and $\ell_{\text{rbf}}$ being the characteristic length scale. We derive the gradients based on this RBF kernel in the following and present other options (e.g., Matérn kernel and Rational Quadratic) in the Appendix and numerical experiments. For any trajectory $\tau = \{(s_0, a_0), (s_1, a_1), \dots, (s_{T-1}, a_{T-1})\}$, the joint ground-truth reward vector follows a multivariate Gaussian:
\begin{align*}
\mathbf{r}_b = [R_b(s_0,a_0), \dots, R_b(s_{T-1},a_{T-1})]^\top \sim \mathcal{N}\big(\boldsymbol{\mu}_{\boldsymbol\theta}, \mathbf{K}_{\boldsymbol\phi}\big),
\end{align*}
where $\boldsymbol{\mu}_{\boldsymbol{\theta}} = [\mu_{\boldsymbol{\theta}}(s_0, a_0), \mu_{\boldsymbol{\theta}}(s_1, a_1), \dots, \mu_{\boldsymbol{\theta}}(s_{T-1}, a_{T-1})]^\top$ and $[\mathbf{K}_{\boldsymbol\phi}]_{ij} = k_{\boldsymbol\phi}((s_i, a_i), (s_j, a_j))$.

To compute the likelihood of such a reward vector, we need a proxy reward for each step. In environments with only episodic reward $R_{ep}(\tau)$, we employ the following LOO strategy to construct training targets. Specifically, for each time step $i$, we define the LOO target as the following:
\begin{align}
\tilde{r}(s_i, a_i) = R_{\mathrm{ep}}(\tau) - \sum_{t=0, t \neq i}^{T-1} \mu_{\boldsymbol\theta}(s_t, a_t).
\end{align}
These LOO targets serve as noisy observations of the ground-truth rewards, forming a training dataset $\mathcal{D}_{\tau} = \{((s_i, a_i), \tilde{r}(s_i, a_i))\}_{i=0}^{T-1}$. Given such a dataset, we optimize a \emph{pseudo-likelihood} that treats the LOO targets as noisy observations:
\begin{align}
p(\tilde{\mathbf{r}} \mid \boldsymbol\theta, \boldsymbol\phi, \sigma_{\epsilon}) 
= \mathcal{N}(\boldsymbol{\mu}_{\boldsymbol\theta}, \mathbf{K}_{\boldsymbol\phi} + \sigma_\epsilon^2\mathbf{I}).
\end{align}
Throughout this section, the LOO targets $\tilde{\mathbf r}$ are constructed at the current $\boldsymbol\theta$ and then treated as constants during gradient updates (i.e., no backpropagation through $\tilde{\mathbf r}$).


To learn the hyperparameters $\boldsymbol\theta, \boldsymbol\phi, \sigma_{\epsilon}$, we maximize the log marginal likelihood of the reward targets on trajectory $\tau$, which corresponds to the following:
\begin{align}
\label{eq:gp-log-likelihood}
\log p(\tilde{\mathbf{r}} \mid \boldsymbol\theta, \boldsymbol\phi, \sigma_{\epsilon}) &= \underbrace{-\frac{1}{2}(\tilde{\mathbf{r}} - \boldsymbol{\mu}_{\boldsymbol\theta})^\top(\mathbf{K}_{\boldsymbol\phi} + \sigma_\epsilon^2\mathbf{I})^{-1}(\tilde{\mathbf{r}} - \boldsymbol{\mu}_{\boldsymbol\theta})}_{\text{data fitting term}} \notag\\
&\hspace{0.3cm} \underbrace{-\frac{1}{2}\log\det(\mathbf{K}_{\boldsymbol\phi} + \sigma_\epsilon^2\mathbf{I})}_{\text{Occam factor}} - \frac{|\tau|}{2}\log(2\pi),
\end{align}
where 
the first term (``data fitting'') measures how well the model explains the data, and the second term (``Occam factor'') penalizes model overfitting. Then, our problem can be formulated as 
\begin{align*}
    \max_{\boldsymbol\theta, \boldsymbol\phi, \sigma_{\epsilon}} \log p(\tilde{\mathbf{r}} \mid \boldsymbol\theta, \boldsymbol\phi, \sigma_{\epsilon}),
\end{align*}
or equivalently,
\begin{align}\label{eq:GP-Obj}
    \min_{\boldsymbol\theta, \boldsymbol\phi, \sigma_{\epsilon}} \mathcal{L}(\tau;\boldsymbol\theta, \boldsymbol\phi, \sigma_{\epsilon})=- \log p(\tilde{\mathbf{r}} \mid \boldsymbol\theta, \boldsymbol\phi, \sigma_{\epsilon}).
\end{align}
To optimize this objective, we compute the gradient with respect to each hyperparameter and update the parameter in the gradient descent direction. The detailed gradient derivations and their theoretical implications are presented in the Theoretical Analyses section.

\subsection{Algorithms}
In this section, we present the algorithm of our GP-LRR. Our approach follows an iterative paradigm that alternates between reward modeling via GP regression and policy optimization via SAC using the learned reward signals.

\begin{algorithm}[h!]
\caption{Gaussian Process Reward Redistribution}
\label{algo:GPRR}
\begin{algorithmic}[1]
\STATE \textbf{Input:} Initialize parameters $\boldsymbol\theta, \sigma_f^2, \ell_{\text{rbf}}, \sigma_{\epsilon}^2$, replay buffer $\mathcal{D}_{\tau}$, batch size $M$, learning rate $\beta$, gradient steps $G$
\STATE Sample trajectory batch $\{(\tau_m, R_{\mathrm{ep}}(\tau_m))\}_{m=1}^{M}$ from $\mathcal{D}_{\tau}$ 
\STATE Initialize $\mathcal{L} = 0$ 
\FOR{$m \in \{1, \dots, M\}$} 
    \STATE Compute mean vector
        \begin{align*}
        \boldsymbol{\mu}_m = [\mu_{\boldsymbol\theta}(s_0^m, a_0^m), \ldots, \mu_{\boldsymbol\theta}(s_{|\tau_m|-1}^m, a_{|\tau_m|-1}^m)]^\top
        \end{align*}
    \STATE Compute kernel {using trajectory $\tau_m$}
        \begin{align*}
        [\mathbf{K}_m]_{ij} = \sigma_f^2 \exp\left(-\frac{\|(s_i^m, a_i^m) - (s_j^m, a_j^m)\|^2}{2\ell_{\text{rbf}}^2}\right)
        \end{align*}
    \FOR{$i \in \{0, \ldots, |\tau_m|-1\}$}
        \STATE $\tilde{r}(s_i^m, a_i^m) = R_{\mathrm{ep}}(\tau_m) - \sum_{t=0, t\neq i}^{|\tau_m|-1} \mu_{\boldsymbol\theta}(s_t^m, a_t^m)$
    \ENDFOR
    \STATE Construct LOO targets $\tilde{\mathbf{r}}_m=[\tilde{r}(s_0^m, a_0^m),\ldots,\tilde{r}(s_{|\tau_m|-1}^m, a_{|\tau_m|-1}^m)]^{\top}$.
    \STATE Compute $\mathbf{K}_{\sigma} = \mathbf{K}_m + \sigma_\epsilon^2\mathbf{I}$
    \STATE Compute $\boldsymbol{\alpha}_m = \mathbf{K}_{\sigma}^{-1}(\tilde{\mathbf{r}}_m - \boldsymbol{\mu}_m)$
    \STATE Compute trajectory negative log marginal likelihood
        \begin{align*}
        \mathcal{L}_m \propto \frac{1}{2}(\tilde{\mathbf{r}}_m - \boldsymbol{\mu}_m)^\top\boldsymbol{\alpha}_m + \frac{1}{2}\log\det{(\mathbf{K}_{\sigma})}
        \end{align*}
    \STATE $\mathcal{L} \leftarrow \mathcal{L} + \mathcal{L}_m / M$
\ENDFOR
\FOR{gradient step $= 1, \dots, G$}
    \STATE Update the hyperparameters
    \begin{align*}
    \boldsymbol\theta &\gets \boldsymbol\theta - \beta\,\nabla_{\boldsymbol\theta} \mathcal{L}, \quad
    \sigma_f^2 \gets \sigma_f^2 - \beta\,\nabla_{\sigma_f^2} \mathcal{L} \\
    \ell_{\text{rbf}} &\gets \ell_{\text{rbf}} - \beta\,\nabla_{\ell_{\text{rbf}}} \mathcal{L}, \quad
    \sigma_\epsilon^2 \gets \sigma_\epsilon^2 - \beta\,\nabla_{\sigma_\epsilon^2} \mathcal{L}
    \end{align*}
\ENDFOR
\end{algorithmic}
\end{algorithm}

Algorithm~\ref{algo:GPRR} details the core GP reward redistribution procedure. For each trajectory batch, we construct LOO targets and optimize the marginal likelihood by performing gradient descent steps. The matrix inverse $\mathbf{K}_{\sigma}^{-1}(\tilde{\mathbf{r}}_j - \boldsymbol{\mu}_j)$ can be computed via Cholesky decomposition for numerical stability and efficiency.

Algorithm~\ref{algo:sac-gprr} then integrates GP-LRR with SAC by using the learned mean function $\mu_{\boldsymbol\theta}(s,a)$ as dense reward signals. We maintain two buffer sets: transition buffer $\mathcal{D}$ for SAC updates and complete trajectory buffer $\mathcal{D}_{\tau}$ for GP training. While GP training has $\mathcal{O}(M|\tau|^3)$ complexity, we amortize this cost by updating the GP model every $n_{\text{update}}$ episodes while performing SAC updates more frequently.

\begin{algorithm}[htbp]
\caption{SAC with GP-LRR}
\label{algo:sac-gprr}
\begin{algorithmic}[1]
\STATE \textbf{Initialize:} Critic networks $Q_{\omega_1}, Q_{\omega_2}$, actor $\pi_\zeta$, GP reward model parameters $\boldsymbol\theta, \sigma_f^2, \ell_{\text{rbf}}, \sigma_\epsilon^2$, Transition buffer $\mathcal{D} \gets \emptyset$, trajectory buffer $\mathcal{D}_\tau \gets \emptyset$
\FOR{episode $= 1, 2, \ldots$}
   \STATE Collect trajectory $\tau = \{(s_t, a_t)\}_{t=0}^{T-1}$ by following $\pi_\zeta$
   \STATE Store trajectory: $\mathcal{D}_\tau \gets \mathcal{D}_\tau \cup \{(\tau, R_{\text{ep}}(\tau))\}$
   \STATE Store transitions: $\mathcal{D} \gets \mathcal{D} \cup \{(s_t, a_t, s_{t+1}) : t \in [0,T-1]\}$
   \IF{episode $\bmod$ $n_{\text{update}} = 0$}
       \STATE Update GP model using Algorithm~\ref{algo:GPRR} with batch from $\mathcal{D}_\tau$
   \ENDIF
   \FOR{gradient step $= 1, \ldots, G$}
       \STATE Sample minibatch $\mathcal{B} = \{(s_i, a_i, s'_i)\}_{i=1}^{B} \sim \mathcal{D}$
       \STATE Compute dense rewards $r_i = \mu_{\boldsymbol\theta}(s_i, a_i)$,\, $i \in [1,B]$
       \STATE Compute TD targets following~\eqref{eq:SAC-target} using $r_i$
       \STATE Update critics $\omega_1, \omega_2$ by minimizing~\eqref{eq:sac_critic}
       \STATE Update actor $\pi_\zeta$ by minimizing~\eqref{eq:sac_actor}
       \STATE Update temperature $\alpha$ by minimizing~\eqref{eq:sac_temp}
   \ENDFOR
\ENDFOR
\end{algorithmic}
\end{algorithm}

\subsection{Theoretical Analyses}
\label{subsec:GPAnalysis}
In this section, we analyze the theoretical properties of GP-LRR. We begin by establishing that our framework generalizes existing MSE-based approaches. Then, we demonstrate how the GP structure enables more effective credit assignment through correlation modeling.

\begin{proposition}[MSE as a Special Case]
\label{prop:GP_to_MSE}
The traditional MSE-based reward redistribution approach emerges as a special case of our GP framework. Specifically, when the kernel matrix reduces to identity ($\mathbf{K}_{\boldsymbol\phi} = \mathbf{I}$) and observation noise vanishes ($\sigma_\epsilon = 0$), the objective function for \eqref{eq:GP-Obj} becomes
\begin{align*}
\mathcal{L}(\tau; \boldsymbol\theta) 
&\propto \frac{|\tau|}{2}\bigg(R_{\text{ep}}(\tau) - \sum_{t=0}^{|\tau|-1}\mu_{\boldsymbol\theta}(s_t, a_t)\bigg)^2 
\end{align*}
\end{proposition}
This proposition shows that MSE objective \eqref{eq:MSERewardRedistribution} corresponds to a degenerate GP with no correlation structure between state-action pairs.

\begin{remark}
MSE-based methods assume independent per-step rewards, causing high variance under sparse feedback. By kernelizing correlations, our GP framework pools information across related state–action pairs, yielding smoother and more sample-efficient credit assignment.
\end{remark}

In GP–LRR, per‑step errors are not treated in isolation but are aggregated through the precision matrix to form a \emph{temporal credit‑assignment network}:
\begin{proposition}[Gradient Flow with Correlations]
\label{prop:GP_gradient}
The gradient of the negative log marginal likelihood $\mathcal{L}(\tau;\boldsymbol\theta, \boldsymbol\phi, \sigma_{\epsilon})$ with respect to the mean function parameters $\boldsymbol\theta$ is
\begin{align}
\frac{\partial \mathcal{L}}{\partial \boldsymbol\theta} = -\underbrace{\frac{\partial \boldsymbol{\mu}_{\boldsymbol\theta}^\top}{\partial \boldsymbol\theta}}_{\text{Neural Network Jacobian}} \hspace{0.1cm} \cdot \hspace{0.1cm} \underbrace{\mathbf{K}_{\sigma}^{-1} (\tilde{\mathbf{r}} - \boldsymbol{\mu}_{\boldsymbol\theta})}_{\text{precision-weighted residual}}
\end{align}
For a specific state-action pair $(s_i, a_i)$, the contribution to its gradient is formed by the prediction errors from all state-action pairs:
\begin{align*}
[\mathbf{K}_{\sigma}^{-1}(\tilde{\mathbf{r}} - \boldsymbol{\mu}_{\boldsymbol\theta})]_i &= w_{ii}(\tilde{r}(s_i, a_i) - \mu_{\boldsymbol\theta}(s_i, a_i)) \notag\\
&\quad + \sum_{j \neq i} w_{ij}(\tilde{r}(s_j, a_j) - \mu_{\boldsymbol\theta}(s_j, a_j)),
\end{align*}
where $w_{ij} = [\mathbf{K}_{\sigma}^{-1}]_{ij}$ is the element of the precision matrix $\mathbf{K}_{\sigma}^{-1}=(\mathbf{K}_{\boldsymbol\phi} + \sigma_\epsilon^2\mathbf{I})^{-1}$.
\end{proposition}

\begin{remark}[Credit Assignment through Correlations]
This decomposition reveals the fundamental difference between GP and MSE approaches. In MSE where $\mathbf{K}_{\sigma} = \mathbf{I}$, we have $w_{ij} = 0$ for $i \neq j$, so each gradient depends only on its own prediction error. In contrast, the GP formulation creates a \emph{credit assignment network} through the precision matrix: when $(s_i, a_i)$ and $(s_j, a_j)$ are highly correlated, their errors are pooled together during learning. This is particularly powerful for temporal credit assignment—errors at future time steps can directly influence the reward estimates at earlier time steps, addressing the fundamental challenge of learning from sparse episodic feedback.
\end{remark}

\begin{proposition}[Length Scale and Smoothness Trade-off]
\label{prop:length_scale}
The gradient of the negative log marginal likelihood $\mathcal{L}(\tau;\boldsymbol\theta, \boldsymbol\phi, \sigma_{\epsilon})$ with respect to the RBF length scale $\ell_{\text{rbf}}$ is
\begin{align}
\frac{\partial \mathcal{L}}{\partial \ell_{\text{rbf}}} = \frac{1}{2}\text{tr}\left(\mathbf{K}_{\sigma}^{-1}\frac{\partial \mathbf{K}_{\boldsymbol\phi}}{\partial \ell_{\text{rbf}}}\right) - \frac{1}{2}\boldsymbol{\alpha}^T \frac{\partial \mathbf{K}_{\boldsymbol\phi}}{\partial \ell_{\text{rbf}}} \boldsymbol{\alpha},
\end{align}
where $tr(A)$ denotes the trace of matrix $A$, $\boldsymbol\alpha = \mathbf{K}_{\sigma}^{-1}(\tilde{\mathbf{r}} - \boldsymbol\mu_{\boldsymbol\theta})$ and $\big[\frac{\partial\mathbf{K}_{\boldsymbol\phi}}{\partial\ell_{\text{rbf}}}\big]_{ij} = k_{ij}\frac{d_{ij}^2}{\ell_{\text{rbf}}^3}$ with {$d_{ij} = \|(s_i, a_i) - (s_j, a_j)\|$, and $k_{ij} = k_{\phi}((s_i, a_i), (s_j, a_j))$.}
\end{proposition}

\begin{remark}[Automatic Smoothness Adaptation]
The gradient in Proposition~\ref{prop:length_scale} reveals how the GP framework automatically balances model complexity and data fitting through the length scale parameter $\ell_{\text{rbf}}$. Here, a smaller  $\ell_{\text{rbf}}$
leads to a more complex model that can fit the data better, while a larger $\ell_{\text{rbf}}$ results in a smoother model that generalizes more. By maximizing marginal likelihood, GP-LRR automatically adjusts $\ell_{\text{rbf}}$, thus balancing data fitting with model complexity.

\end{remark}

\begin{proposition}[Observation Noise Adaptation]
\label{prop:noise_adaptation}
The gradient of the negative log marginal likelihood $\mathcal{L}(\tau;\boldsymbol\theta, \boldsymbol\phi, \sigma_{\epsilon})$ with respect to the observation noise variance $\sigma_{\epsilon}^2$ is:
\begin{align}
\frac{\partial \mathcal{L}}{\partial \sigma_\epsilon^2} = \frac{1}{2}\text{tr}(\mathbf{K}_{\sigma}^{-1}) - \frac{1}{2}\|\boldsymbol{\alpha}\|^2,
\end{align}
where $\boldsymbol\alpha = \mathbf{K}_{\sigma}^{-1}(\tilde{\mathbf{r}} - \boldsymbol\mu_{\boldsymbol\theta})$.
\end{proposition}

\begin{remark}[Automatic Noise Level Selection]
The trace term $\text{tr}(\mathbf{K}_{\sigma}^{-1})$ increases as $\sigma_\epsilon^2$ decreases, acting as a regularizer that prevents the noise from vanishing. Note that as $\sigma_\epsilon^2 \to 0$, we have $\mathbf{K}_{\sigma}^{-1} \approx \mathbf{K}_{\boldsymbol\phi}^{-1}$, and the trace can become arbitrarily large for ill-conditioned kernels. The second term $\|\boldsymbol{\alpha}\|^2 = (\tilde{\mathbf{r}} - \boldsymbol{\mu}_{\boldsymbol\theta})^T\mathbf{K}_{\sigma}^{-2}(\tilde{\mathbf{r}} - \boldsymbol{\mu}_{\boldsymbol\theta})$ measures the weighted prediction error---when the model cannot explain the LOO targets well, this term becomes large and drives $\sigma_\epsilon^2$ upward to accommodate the misfit.
\end{remark}

\section{Experiments}
In this section, we evaluate our proposed GP-LRR framework on continuous control tasks with delayed episodic rewards. We compare GP-LRR against several baselines across multiple MuJoCo environments to demonstrate its effectiveness in credit assignment and sample efficiency. All experiments were conducted on a single NVIDIA RTX 4080 Super GPU with 16GB memory.

\subsection{Experimental Setup}
We evaluate our algorithm on four continuous control tasks: HalfCheetah-v4, Hopper-v4, Swimmer-v4, and Walker2d-v4. Following prior work~\citep{ren2021learning}, we modify these environments to provide only episodic rewards, where the agent receives zero reward at all intermediate steps and only observes the cumulative return $R_{\text{ep}}(\tau) = \sum_{t=0}^{T-1} r_t$ at episode termination, which creates a challenging sparse reward setting that tests each method's ability to perform temporal credit assignment.

\begin{figure}[h!]
\centering
\begin{tabular}{cc}
\includegraphics[width=0.45\linewidth]{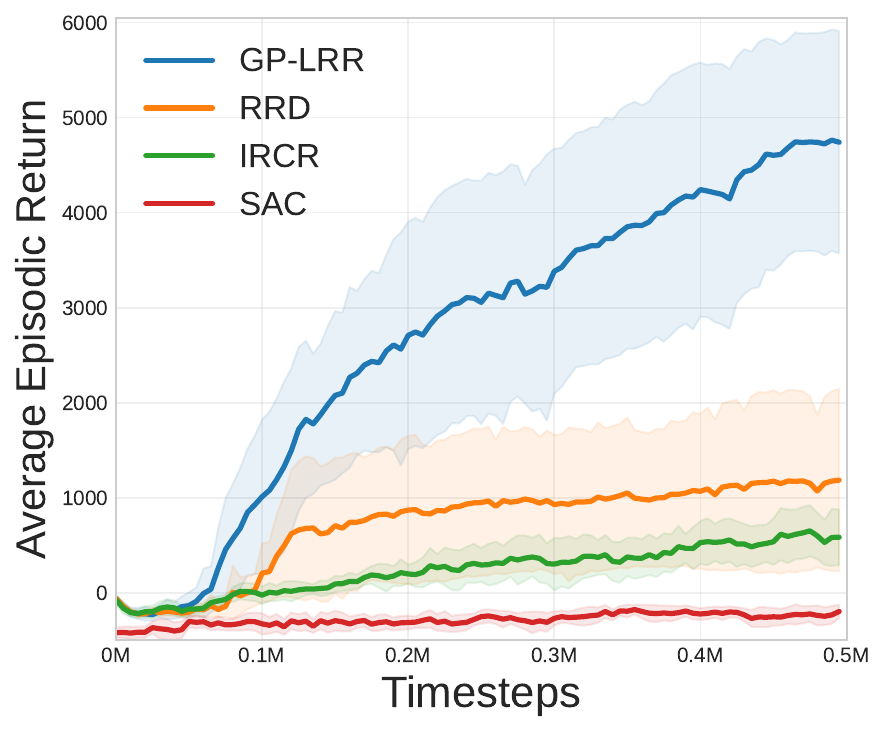} & 
\includegraphics[width=0.45\linewidth]{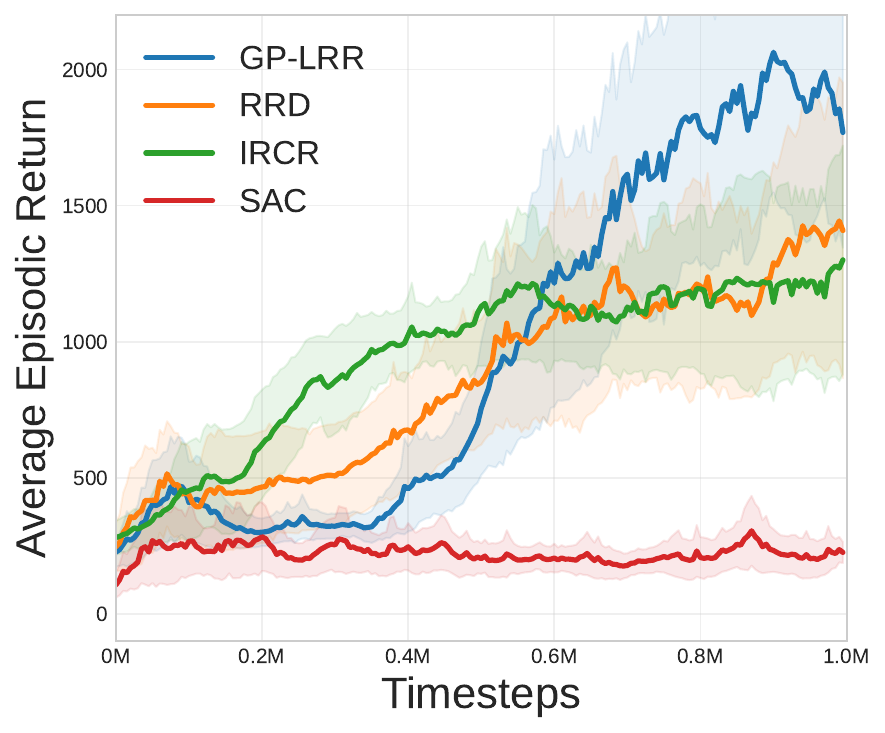}  \\[-0.1cm]
{\small (a) HalfCheetah-v4} & {\small (b) Hopper-v4} \\[0.2cm]
\includegraphics[width=0.45\linewidth]{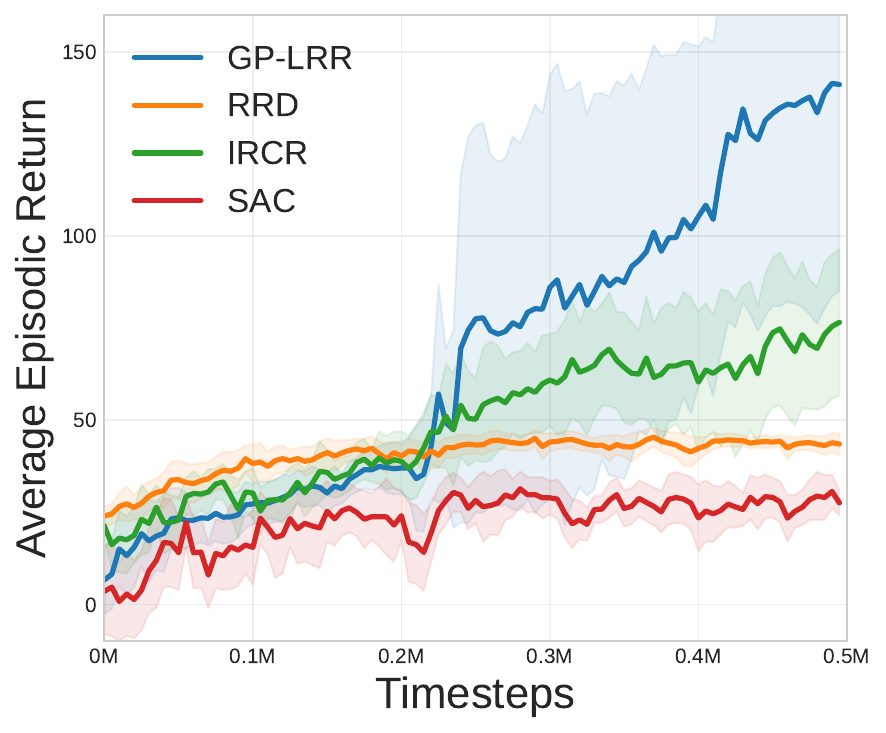} &
\includegraphics[width=0.45\linewidth]{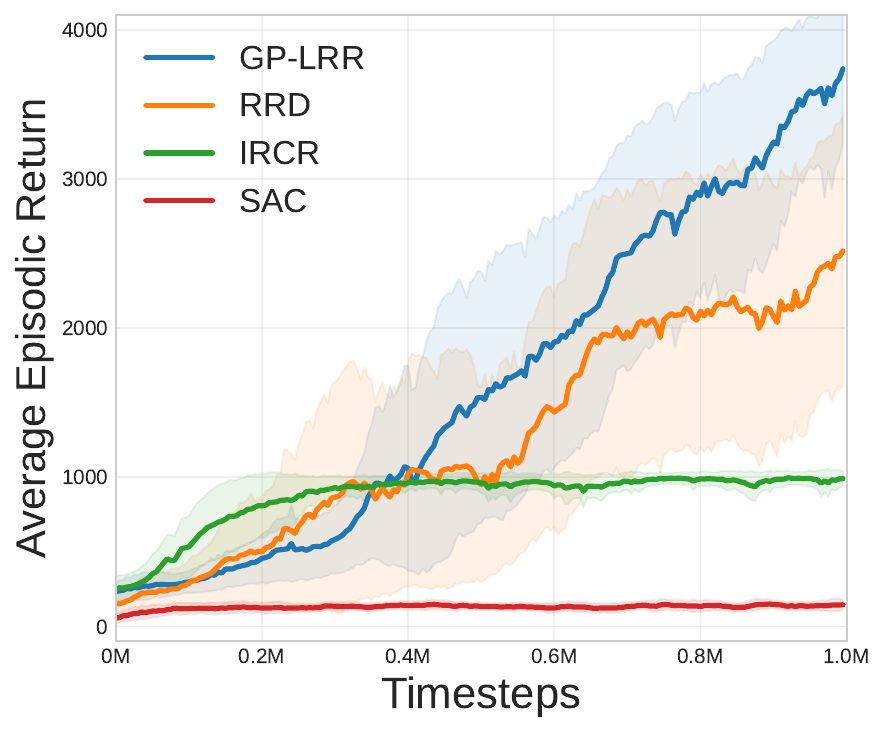} \\[-0.1cm]
{\small (c) Swimmer-v4} & {\small (d) Walker2d-v4}
\end{tabular}
\caption{Learning curves on different MuJoCo environments. Solid curves show average returns over 5 independent runs; shaded regions indicate standard deviation. Performance is evaluated every 5,000 environment steps. GP-LRR uses the RBF kernel in these experiments.}
\label{fig:comparison}
\end{figure}

We compare GP-LRR against three baselines. First, we evaluate vanilla Soft Actor-Critic (SAC)~\citep{haarnoja2018soft} using only the sparse episodic reward signal, which serves as a lower bound representing performance without any reward redistribution. Second, we test IRCR~\citep{gangwani2020learning}, which uniformly redistributes the episodic return across all timesteps by assigning $r_t = R_{\text{ep}}(\tau) / T$ to each state-action pair. Third, we compare against RRD~\citep{ren2021learning}, the state-of-the-art randomized return decomposition method that approximates full trajectory decomposition through random subsequence sampling. We use RRD's recommended configuration with sampling parameter $K=64$, which achieved the best performance in their experiments. Note that we do not compare GP-LRR with RUDDER, as RRD has been shown to consistently outperform RUDDER in  \citep{ren2021learning}.

All methods use SAC as the base RL algorithm with identical hyperparameters. For GP-LRR, the kernel hyperparameters including length scale $\ell_{\text{rbf}}$, signal variance $\sigma_f^2$, and noise variance $\sigma_\epsilon^2$ are learned jointly via gradient descent on the marginal likelihood. We update the GP reward model every 100 environment steps using trajectory batches of size 4.

\subsection{Comparison with Baselines}
Figure~\ref{fig:comparison} presents the learning curves comparing GP-LRR against the three baselines across four representative environments.\footnote{Note that we only evaluate these four environments since not all MuJoCo environments are suitable for modeling state-action pair correlations.} We report the mean and standard deviation with random initialization over 5 independent simulation runs. For GP-LRR method, we adopted the RBF kernel.

GP-LRR consistently outperforms all baselines across the evaluated environments, with particularly striking improvements in HalfCheetah-v4 where the smooth dynamics and spatially correlated rewards align well with the GP's principled correlation modeling. Unlike methods requiring on-policy samples~\citep{guo2018generative, zheng2018learning}, GP-LRR seamlessly integrates with off-policy algorithms like SAC, further improving sample efficiency.

\subsection{Ablation Study}
To understand the key components of GP-LRR, we conduct ablation studies examining the effect of kernel choice and length scale initialization on performance.

\subsubsection{Effect of Kernel Choice}
In GP‑LRR, the kernel choice shapes reward interpolation: (i) the RBF kernel yields infinitely smooth, Gaussian‑weighted correlations ideal for gently varying reward landscapes, (ii) the Matérn–3/2 kernel produces once‑differentiable, rougher paths that handle abrupt changes, and (iii) the Rational Quadratic kernel—an infinite mixture of RBFs with heavy‑tailed scales—adaptively captures both fine details and broad trends.

\begin{figure}[h!]
\centering
\begin{tabular}{cc}
\includegraphics[width=0.45\linewidth]{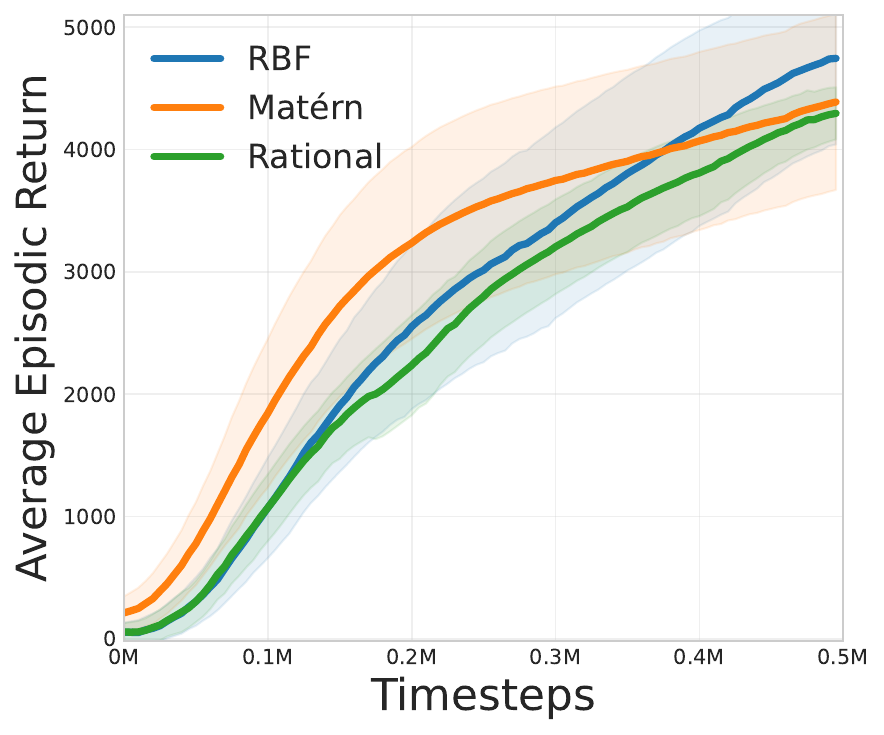} & 
\includegraphics[width=0.45\linewidth]{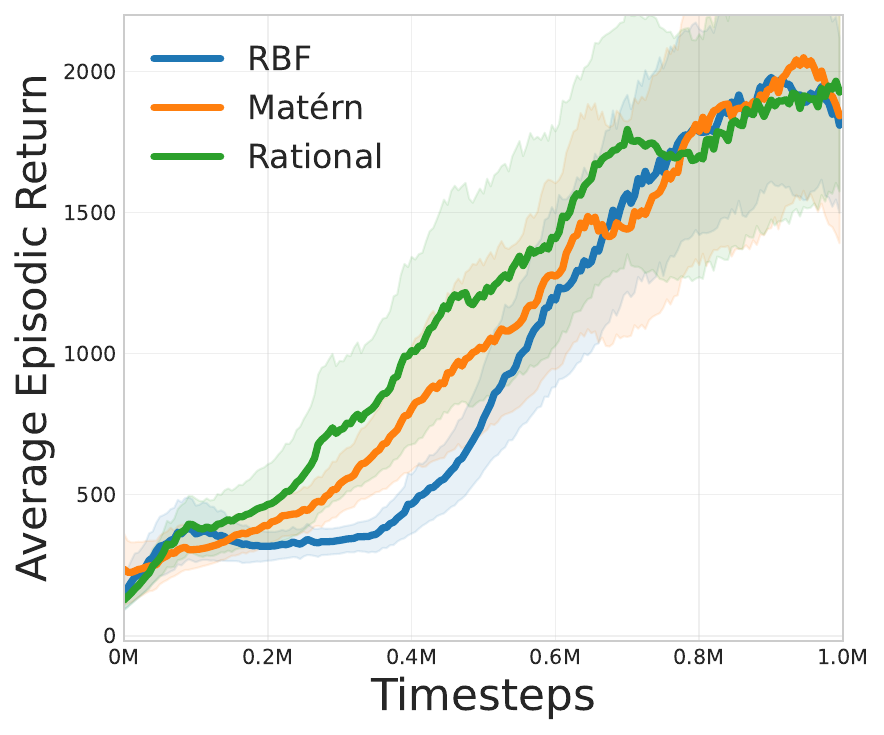} \\[-0.1cm]
{\small (a) HalfCheetah-v4} & {\small (b) Hopper-v4} \\[0.2cm]
\includegraphics[width=0.45\linewidth]{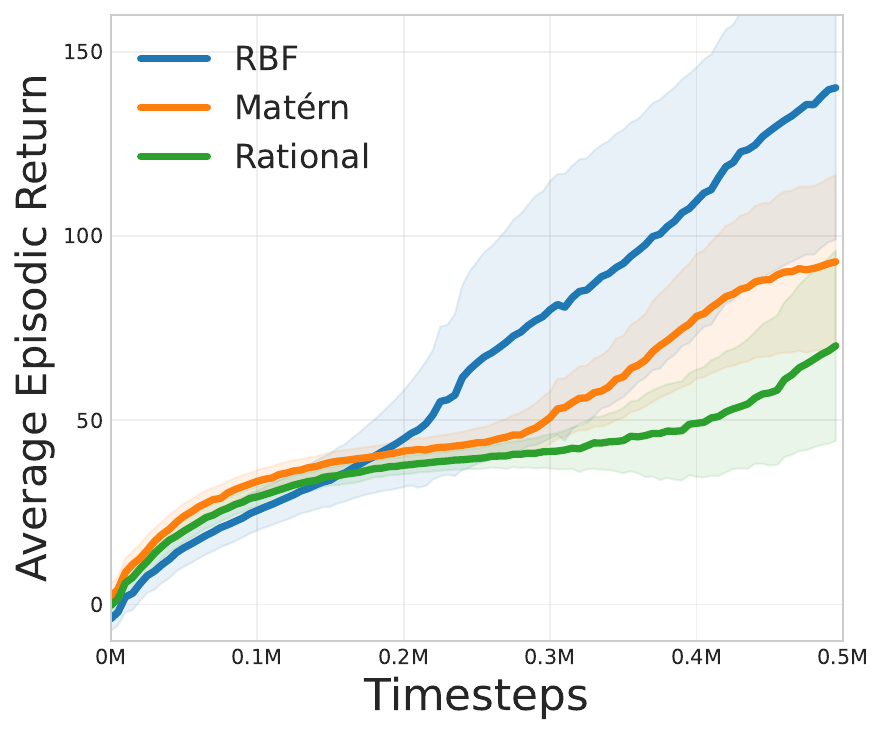} &
\includegraphics[width=0.45\linewidth]{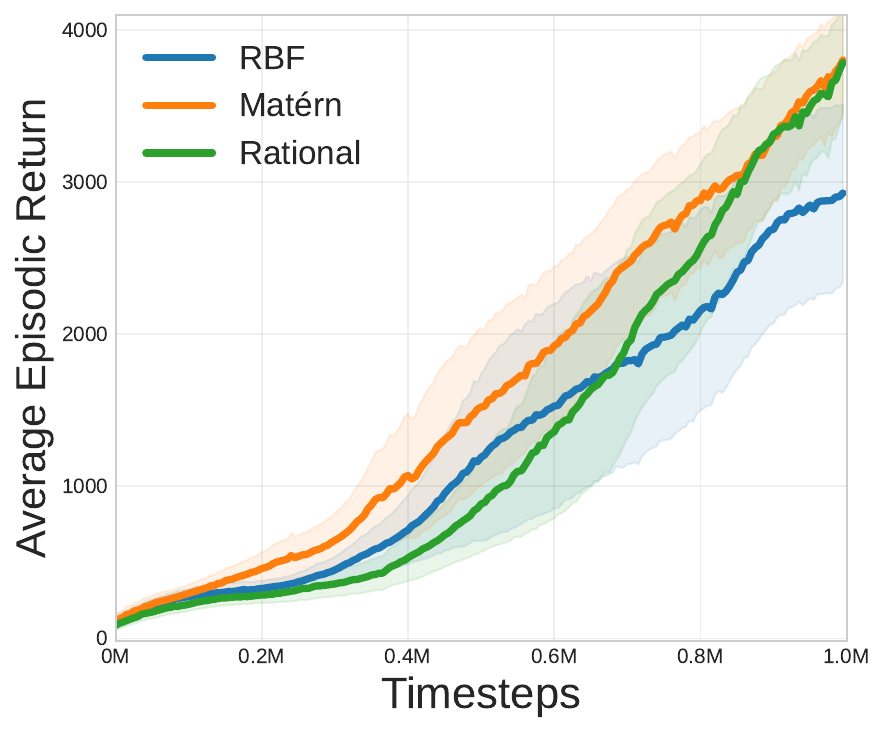} \\[-0.1cm]
{\small (c) Swimmer-v4} & {\small (d) Walker2d-v4}
\end{tabular}
\caption{Learning curves on different MuJoCo environments with sparse episodic rewards using GP-LRR method with various kernels. Solid curves show average returns over 5 independent runs; shaded regions indicate standard deviation. Performance is evaluated every 5,000 environment steps.}
\label{fig:kernel}
\end{figure}

The results in Figure~\ref{fig:kernel} reveal that kernel performance is task-dependent. While RBF excels in HalfCheetah, Hopper, and Swimmer—environments with smooth reward landscapes—it performs worst in Walker2d, where Matérn and Rational Quadratic kernels achieve superior results. This suggests that Walker2d's reward structure may contain discontinuities or sharp transitions that the infinitely differentiable RBF kernel cannot capture effectively. The Matérn kernel's ability to model less smooth functions proves advantageous in this more complex balancing task.

\subsubsection{Impact of Length Scale Initialization}

The length scale parameter $\ell_{\text{rbf}}$ controls how far correlations extend in the state-action space. Small values create localized credit assignment where only nearby states influence each other, while large values enable global information sharing across distant states. This parameter essentially determines whether the GP assumes rewards change rapidly (small $\ell_{\text{rbf}}$) or smoothly (large $\ell_{\text{rbf}}$) across the state space.

We test four \textit{initialization} strategies: small ($\ell = 0.1$) for environments with sharp reward transitions, medium ($\ell = 1.0$) as a balanced default, large ($\ell = 10.0$) for smooth reward landscapes, and adaptive initialization based on median pairwise distances in early trajectories.

\begin{figure}[h!]
\centering
\begin{tabular}{cc}
\includegraphics[width=0.45\linewidth]{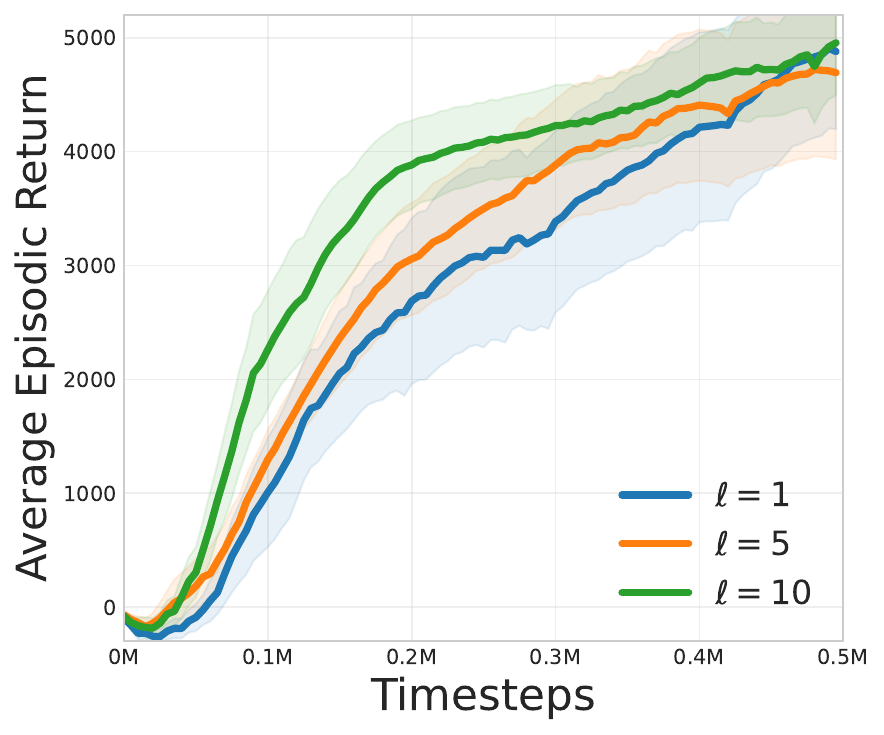} & 
\includegraphics[width=0.45\linewidth]{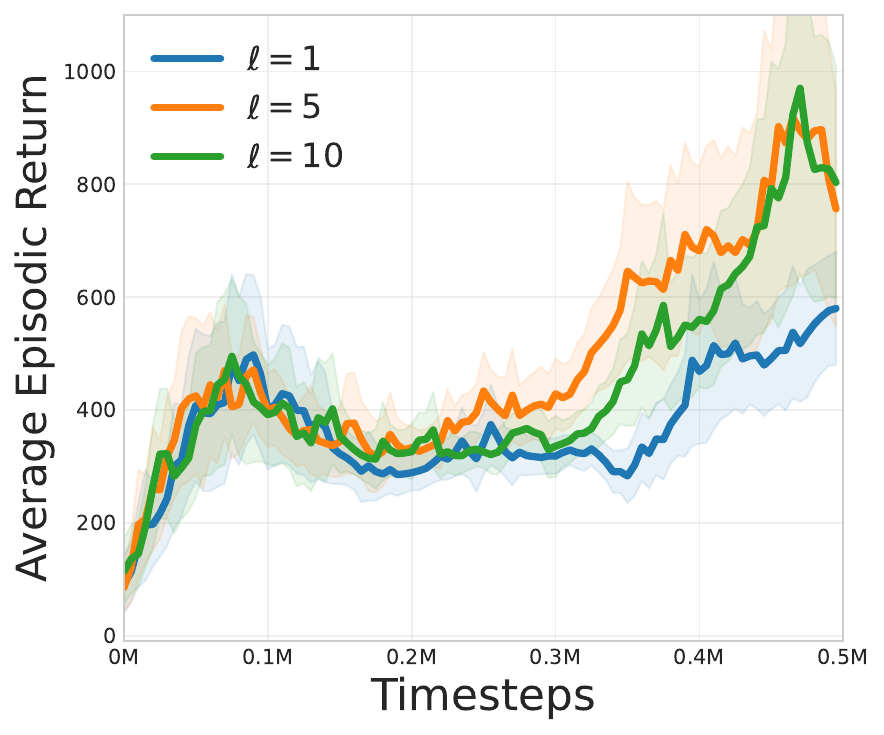} \\[-0.1cm]
{\small (a) HalfCheetah-v4} & {\small (b) Hopper-v4} \\[0.2cm]
\includegraphics[width=0.45\linewidth]{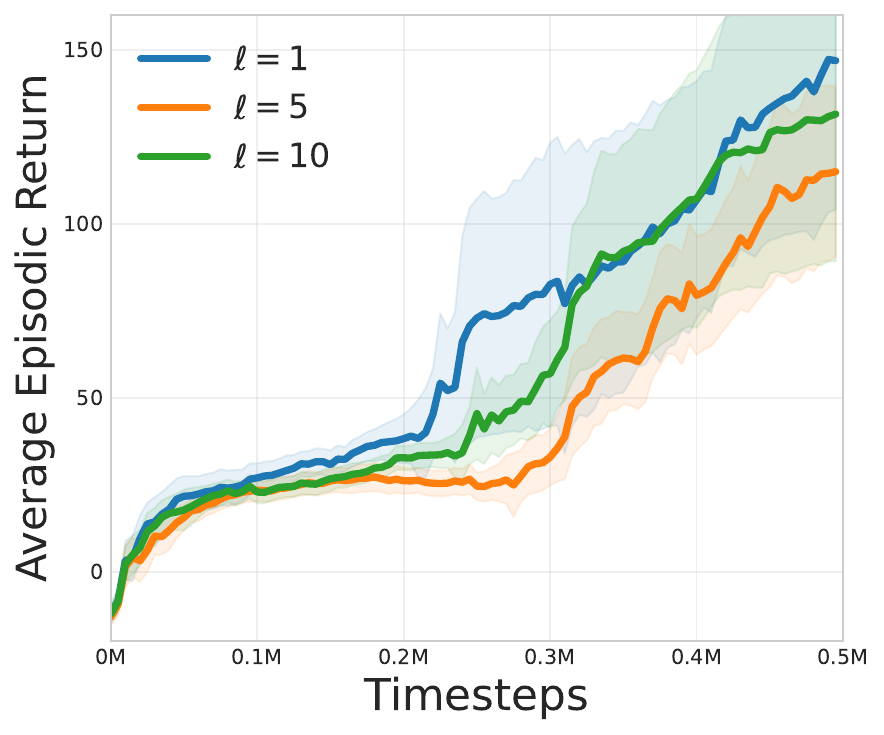} &
\includegraphics[width=0.45\linewidth]{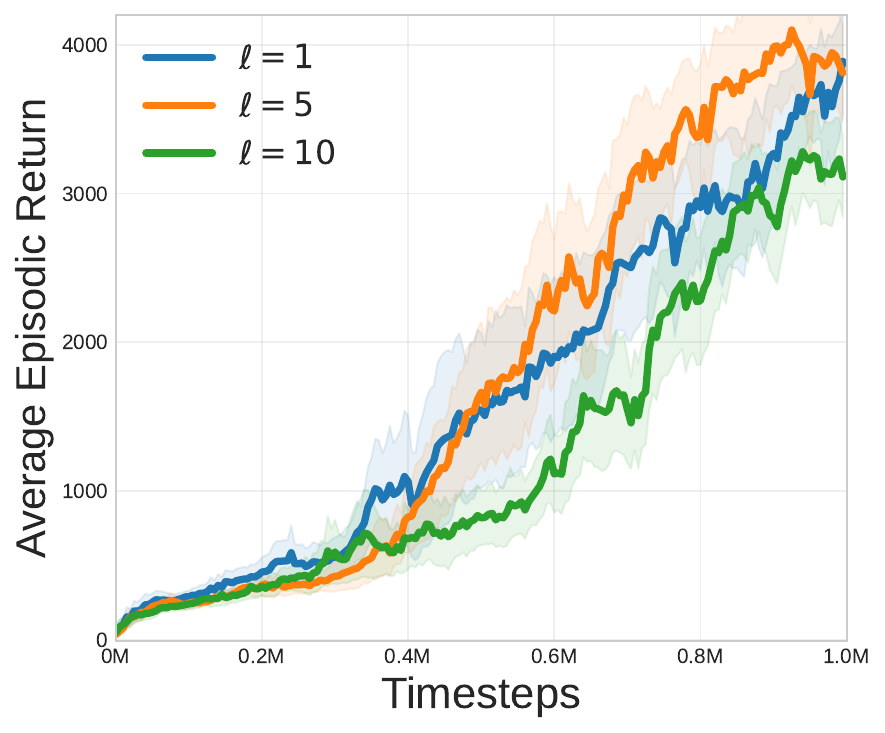}  \\[-0.1cm]
{\small (c) Swimmer-v4} & {\small (d) Walker2d-v4}
\end{tabular}
\caption{Learning curves on different MuJoCo environments with sparse episodic rewards using GP-LRR method with different $\ell_{\text{rbf}}$ initialization. Solid curves show average returns over 5 independent runs; shaded regions indicate standard deviation. Performance is evaluated every 5,000 environment steps.}
\label{fig:ell}
\end{figure}

The results in Figure~\ref{fig:ell} show no universal optimal length scale initialization—each environment responds differently. HalfCheetah prefers large values ($\ell = 10$), Walker2d favors medium scales, while Hopper and Swimmer show high variance or insensitivity to initialization. This environment-specific behavior confirms that correlation structure depends heavily on task dynamics, with no one-size-fits-all solution despite GP-LRR's adaptive optimization.

\section{Conclusion}
We introduced a novel likelihood-based framework (GP-LRR) for reward redistribution in RL with episodic returns, where we modeled the per-step reward as a sample from a GP and maximized the likelihood of a trajectory via an LOO strategy. Unlike other conventional reward decomposition methods, our GP-LRR explicitly incorporates the interdependencies among state-action pairs using the kernel function. Our theoretical analyses illustrated that GP-LRR includes the conventional MSE-based approach as a special case. We further showed that it pools the gradient information from different state-action pairs together and implicitly incorporates an uncertainty regularizer to ensure more efficient parameter updating. Empirical results on MuJoCo benchmarks demonstrated that, when combined with SAC, our method yields dense, robust reward signals that improve both sample efficiency and final policy performance. In future work, we plan to extend this framework to non-Gaussian noise models and investigate integrations with other off-policy algorithms.

\section*{Acknowledgments}
This work was supported in part by the National Science Foundation under Grant \#2331782.

\bibliographystyle{plainnat}
\bibliography{ref}

\newpage
\appendix
\section{Appendix}
\subsection{Numerical Experiments Hyperparameters Setup}
In the following table, we present the hyperparameters used in MuJoCo benchmarks in our numerical section:

\begin{table*}[h!]
\centering
\begin{threeparttable}
\caption{Hyperparameter configuration in MuJoCo environments.}
\label{tab:hyperparams}
\begin{tabular}{lc}
\toprule
\textbf{Hyperparameter} & \textbf{Default Configuration} \\
\midrule
SAC learning rate & $3 \times 10^{-4}$ \\
Reward model learning rate & $1 \times 10^{-3}$ \\
SAC optimizer (all losses) & Adam~\citep{kingma2014adam} \\
Reward model optimizer & Adam~\citep{kingma2014adam} \\
Discount factor $\gamma$ & 0.99 \\
Reward model hidden layers & 2 \\
Reward model neurons per layer & 64 or 256\tnote{*} \\
SAC gradient steps per environment step & 1 \\
Reward model gradient steps per environment step & 100 \\
Transitions in SAC replay buffer & $10^5$ \\
Transitions in reward model replay buffer & 200 \\
Transitions in mini-batch for training SAC & 64 \\
Transitions in mini-batch for training reward model & 4 \\
\bottomrule
\end{tabular}
\begin{tablenotes}
\small
\item[*] For simple environments, a smaller number of neurons (64) is used.
\end{tablenotes}
\end{threeparttable}
\end{table*}

\subsection{Proofs in Section: GP-based Likelihood Reward Redistribution}
\begin{repproposition}{prop:GP_to_MSE}
The traditional MSE-based reward redistribution approach emerges as a special case of our GP framework. Specifically, when the kernel matrix reduces to the identity ($\mathbf{K}_{\boldsymbol\phi}=\mathbf{I}$) and observation noise vanishes ($\sigma_\epsilon\to 0$), the negative log marginal likelihood~\eqref{eq:gp-log-likelihood} becomes
\begin{align*}
\mathcal{L}(\tau;\boldsymbol\theta)\;\propto\;\frac{1}{2}\sum_{i=0}^{|\tau|-1}\Bigl(\tilde r(s_i,a_i)-\mu_{\boldsymbol\theta}(s_i,a_i)\Bigr)^2
\;\propto\;\frac{|\tau|}{2}\Bigl(R_{\mathrm{ep}}(\tau)-\sum_{t=0}^{|\tau|-1}\mu_{\boldsymbol\theta}(s_t,a_t)\Bigr)^2.
\end{align*}
\end{repproposition}

\begin{proof}
Under $\mathbf{K}_{\boldsymbol\phi}=\mathbf{I}$ and $\sigma_\epsilon\to 0$, we have $\mathbf{K}_\sigma=\mathbf{I}$ and hence
\begin{align*}
\mathcal{L}(\tau;\boldsymbol\theta)
= \frac{1}{2}\bigl(\tilde{\mathbf r}-\boldsymbol\mu_{\boldsymbol\theta}\bigr)^\top\bigl(\tilde{\mathbf r}-\boldsymbol\mu_{\boldsymbol\theta}\bigr) + \frac{1}{2}\log|\mathbf{I}| + \frac{|\tau|}{2}\log(2\pi)
= \frac{1}{2}\sum_{i=0}^{|\tau|-1}\bigl(\tilde r(s_i,a_i)-\mu_{\boldsymbol\theta}(s_i,a_i)\bigr)^2 + \text{const}.
\end{align*}
Recall the leave-one-out (LOO) targets are defined \emph{pointwise at the same parameter value $\boldsymbol\theta$} by
\begin{align*}
\tilde r(s_i,a_i) \;=\; R_{\mathrm{ep}}(\tau)\;-\!\!\sum_{\substack{t=0\\ t\neq i}}^{|\tau|-1}\mu_{\boldsymbol\theta}(s_t,a_t).
\end{align*}
At this fixed $\boldsymbol\theta$, the following \emph{algebraic identity} holds for every $i$:
\begin{align*}
\tilde r(s_i,a_i)-\mu_{\boldsymbol\theta}(s_i,a_i)
= R_{\mathrm{ep}}(\tau) - \sum_{t=0}^{|\tau|-1}\mu_{\boldsymbol\theta}(s_t,a_t),
\end{align*}
which is independent of $i$. Therefore,
\begin{align*}
\sum_{i=0}^{|\tau|-1}\bigl(\tilde r(s_i,a_i)-\mu_{\boldsymbol\theta}(s_i,a_i)\bigr)^2
= |\tau|\;\Bigl(R_{\mathrm{ep}}(\tau)-\sum_{t=0}^{|\tau|-1}\mu_{\boldsymbol\theta}(s_t,a_t)\Bigr)^2,
\end{align*}
and the claim follows up to an additive constant and a positive scalar factor $|\tau|/2$.
\end{proof}

\begin{repproposition}{prop:GP_gradient}
The gradient of the negative log marginal likelihood with respect to the mean function parameters $\boldsymbol\theta$ is
\begin{align*}
\frac{\partial \mathcal{L}}{\partial \boldsymbol\theta}
= -\underbrace{\frac{\partial \boldsymbol{\mu}_{\boldsymbol\theta}^\top}{\partial \boldsymbol\theta}}_{\text{Neural Network Jacobian}}
\;\cdot\;
\underbrace{\mathbf{K}_{\sigma}^{-1} (\tilde{\mathbf{r}} - \boldsymbol{\mu}_{\boldsymbol\theta})}_{\text{precision-weighted residual}}.
\end{align*}
For a specific state–action pair $(s_i,a_i)$, the error signal on $\mu_{\boldsymbol\theta}(s_i,a_i)$ aggregates all prediction errors as
\begin{align*}
[\mathbf{K}_{\sigma}^{-1}(\tilde{\mathbf{r}} - \boldsymbol{\mu}_{\boldsymbol\theta})]_i
= w_{ii}\bigl(\tilde{r}(s_i,a_i) - \mu_{\boldsymbol\theta}(s_i,a_i)\bigr)
\;+\; \sum_{j \neq i} w_{ij}\bigl(\tilde{r}(s_j,a_j) - \mu_{\boldsymbol\theta}(s_j,a_j)\bigr),
\end{align*}
where $w_{ij} = [\mathbf{K}_{\sigma}^{-1}]_{ij}$ are the elements of the precision matrix.
\end{repproposition}

\begin{proof}
Throughout this proof, the LOO targets $\tilde{\mathbf r}$ are held fixed (no gradient flows through $\tilde{\mathbf r}$), and $\mathbf K_\sigma$ depends only on $(\boldsymbol\phi,\sigma_\epsilon)$, not on $\boldsymbol\theta$.
The negative log marginal likelihood is
\begin{align*}
\mathcal{L}(\boldsymbol\theta, \boldsymbol\phi, \sigma_\epsilon)
= \frac{1}{2}(\tilde{\mathbf{r}} - \boldsymbol{\mu}_{\boldsymbol\theta})^\top\mathbf{K}_{\sigma}^{-1}(\tilde{\mathbf{r}} - \boldsymbol{\mu}_{\boldsymbol\theta})
+ \frac{1}{2}\log|\mathbf{K}_{\sigma}| + \frac{|\tau|}{2}\log(2\pi).
\end{align*}
Since $\mathbf{K}_{\sigma}$ does not depend on $\boldsymbol\theta$, only the quadratic term depends on $\boldsymbol\theta$:
\begin{align*}
\frac{\partial \mathcal{L}}{\partial \boldsymbol\theta}
&= \frac{\partial}{\partial \boldsymbol\theta}\!\left[\frac{1}{2}(\tilde{\mathbf{r}} - \boldsymbol{\mu}_{\boldsymbol\theta})^\top\mathbf{K}_{\sigma}^{-1}(\tilde{\mathbf{r}} - \boldsymbol{\mu}_{\boldsymbol\theta})\right] \\
&= -\frac{\partial \boldsymbol{\mu}_{\boldsymbol\theta}^\top}{\partial \boldsymbol\theta}\,\mathbf{K}_{\sigma}^{-1}\,(\tilde{\mathbf{r}} - \boldsymbol{\mu}_{\boldsymbol\theta}),
\end{align*}
which yields the stated result. Writing $w_{ij} = [\mathbf{K}_{\sigma}^{-1}]_{ij}$ gives the componentwise form immediately.
\end{proof}

\begin{repproposition}{prop:length_scale}
The gradient of the negative log marginal likelihood with respect to the RBF length scale $\ell_{\text{rbf}}$ is
\begin{align*}
\frac{\partial \mathcal{L}}{\partial \ell_{\text{rbf}}}
= \frac{1}{2}\operatorname{tr}\!\left(\mathbf{K}_{\sigma}^{-1}\frac{\partial \mathbf{K}_{\boldsymbol\phi}}{\partial \ell_{\text{rbf}}}\right)
- \frac{1}{2}\,\boldsymbol{\alpha}^\top \frac{\partial \mathbf{K}_{\boldsymbol\phi}}{\partial \ell_{\text{rbf}}}\,\boldsymbol{\alpha},
\end{align*}
where $\boldsymbol\alpha = \mathbf{K}_{\sigma}^{-1}(\tilde{\mathbf{r}} - \boldsymbol\mu_{\boldsymbol\theta})$ and
$\big[\frac{\partial\mathbf{K}_{\boldsymbol\phi}}{\partial\ell_{\text{rbf}}}\big]_{ij} = k_{ij}\,\frac{d_{ij}^2}{\ell_{\text{rbf}}^3}$ with $d_{ij} = \|(s_i, a_i) - (s_j, a_j)\|_2$ and $k_{ij}=\sigma_f^2\exp(-d_{ij}^2/(2\ell_{\text{rbf}}^2))$.
\end{repproposition}

\begin{proof}
Throughout, the LOO targets $\tilde{\mathbf r}$ are held fixed (no gradient flows through $\tilde{\mathbf r}$), and $\boldsymbol\mu_{\boldsymbol\theta}$ does not depend on $\ell_{\text{rbf}}$. From the negative log marginal likelihood,
\begin{align*}
\mathcal{L}
= \frac{1}{2}(\tilde{\mathbf{r}} - \boldsymbol{\mu}_{\boldsymbol\theta})^\top\mathbf{K}_{\sigma}^{-1}(\tilde{\mathbf{r}} - \boldsymbol{\mu}_{\boldsymbol\theta})
+ \frac{1}{2}\log|\mathbf{K}_{\sigma}| + \frac{|\tau|}{2}\log(2\pi).
\end{align*}
Since $\mathbf{K}_{\sigma} = \mathbf{K}_{\boldsymbol\phi} + \sigma_\epsilon^2\mathbf{I}$ and only $\mathbf{K}_{\boldsymbol\phi}$ depends on $\ell_{\text{rbf}}$, we have
$\frac{\partial\mathbf{K}_{\sigma}}{\partial\ell_{\text{rbf}}} = \frac{\partial\mathbf{K}_{\boldsymbol\phi}}{\partial\ell_{\text{rbf}}}$.
Let $\mathbf v=\tilde{\mathbf r}-\boldsymbol\mu_{\boldsymbol\theta}$. Using
$\frac{\partial}{\partial x}\big(\tfrac12\mathbf a^\top\mathbf A^{-1}\mathbf a\big)
= -\tfrac12 \mathbf a^\top\mathbf A^{-1}\tfrac{\partial \mathbf A}{\partial x}\mathbf A^{-1}\mathbf a$
and
$\frac{\partial}{\partial x}\big(\tfrac12\log|\mathbf A|\big)=\tfrac12\operatorname{tr}(\mathbf A^{-1}\tfrac{\partial \mathbf A}{\partial x})$,
we obtain
\begin{align*}
\frac{\partial \mathcal{L}}{\partial \ell_{\text{rbf}}}
= -\frac{1}{2}\mathbf v^\top\mathbf K_\sigma^{-1}\frac{\partial\mathbf K_{\boldsymbol\phi}}{\partial\ell_{\text{rbf}}}\mathbf K_\sigma^{-1}\mathbf v
+ \frac{1}{2}\operatorname{tr}\!\left(\mathbf K_\sigma^{-1}\frac{\partial\mathbf K_{\boldsymbol\phi}}{\partial\ell_{\text{rbf}}}\right).
\end{align*}
Writing $\boldsymbol\alpha=\mathbf K_\sigma^{-1}\mathbf v$ yields the stated result.
For the RBF kernel $k_{ij} = \sigma_f^2\exp\!\left(-\tfrac{d_{ij}^2}{2\ell_{\text{rbf}}^2}\right)$,
\[
\frac{\partial k_{ij}}{\partial\ell_{\text{rbf}}}
= k_{ij}\,\frac{d_{ij}^2}{\ell_{\text{rbf}}^3},
\]
so $[\partial\mathbf K_{\boldsymbol\phi}/\partial\ell_{\text{rbf}}]_{ij} = k_{ij}\,d_{ij}^2/\ell_{\text{rbf}}^3$.
\end{proof}

\begin{repproposition}{prop:noise_adaptation}
The gradient of the negative log marginal likelihood with respect to the observation noise variance $\sigma_\epsilon^2$ is:
\begin{align*}
\frac{\partial \mathcal{L}}{\partial \sigma_\epsilon^2}
= \frac{1}{2}\operatorname{tr}(\mathbf{K}_{\sigma}^{-1}) - \frac{1}{2}\|\boldsymbol{\alpha}\|^2,
\end{align*}
where $\boldsymbol\alpha = \mathbf{K}_{\sigma}^{-1}(\tilde{\mathbf{r}} - \boldsymbol\mu_{\boldsymbol\theta})$.
\end{repproposition}

\begin{proof}
Throughout, the LOO targets $\tilde{\mathbf r}$ are held fixed (no gradient flows through $\tilde{\mathbf r}$). From the negative log marginal likelihood:
\begin{align*}
\mathcal{L}
= \frac{1}{2}(\tilde{\mathbf{r}} - \boldsymbol{\mu}_{\boldsymbol\theta})^\top\mathbf{K}_{\sigma}^{-1}(\tilde{\mathbf{r}} - \boldsymbol{\mu}_{\boldsymbol\theta})
+ \frac{1}{2}\log|\mathbf{K}_{\sigma}| + \frac{|\tau|}{2}\log(2\pi).
\end{align*}
Since $\mathbf{K}_{\sigma} = \mathbf{K}_{\boldsymbol\phi} + \sigma_\epsilon^2\mathbf{I}$, we have
$\frac{\partial\mathbf{K}_{\sigma}}{\partial\sigma_\epsilon^2} = \mathbf{I}$.
Let $\mathbf v = \tilde{\mathbf r} - \boldsymbol\mu_{\boldsymbol\theta}$. Using the matrix identities
\[
\frac{\partial}{\partial x}\Big(\tfrac12\,\mathbf a^\top\mathbf A^{-1}\mathbf a\Big)
= -\tfrac12\,\mathbf a^\top\mathbf A^{-1}\tfrac{\partial\mathbf A}{\partial x}\mathbf A^{-1}\mathbf a,
\qquad
\frac{\partial}{\partial x}\Big(\tfrac12\log|\mathbf A|\Big)
= \tfrac12\,\operatorname{tr}\!\Big(\mathbf A^{-1}\tfrac{\partial\mathbf A}{\partial x}\Big),
\]
we obtain
\begin{align*}
\frac{\partial \mathcal{L}}{\partial \sigma_\epsilon^2}
&= -\frac{1}{2}\mathbf v^\top\mathbf K_\sigma^{-1}\Big(\tfrac{\partial\mathbf K_\sigma}{\partial\sigma_\epsilon^2}\Big)\mathbf K_\sigma^{-1}\mathbf v
+ \frac{1}{2}\operatorname{tr}\!\Big(\mathbf K_\sigma^{-1}\tfrac{\partial\mathbf K_\sigma}{\partial\sigma_\epsilon^2}\Big) \\
&= -\frac{1}{2}\mathbf v^\top\mathbf K_\sigma^{-2}\mathbf v + \frac{1}{2}\operatorname{tr}(\mathbf K_\sigma^{-1})
= -\frac{1}{2}\|\boldsymbol\alpha\|^2 + \frac{1}{2}\operatorname{tr}(\mathbf K_\sigma^{-1}),
\end{align*}
where $\boldsymbol\alpha=\mathbf K_\sigma^{-1}\mathbf v$. This proves the claim.
\end{proof}

\subsection{Additional Kernel: Matérn 3/2 Kernel Gradients}
We now derive the gradients of the negative log marginal likelihood w.r.t. the Matérn-$3/2$ kernel hyperparameters. Throughout, the LOO targets $\tilde{\mathbf r}$ are held fixed (no gradient flows through $\tilde{\mathbf r}$), and $\boldsymbol\mu_{\boldsymbol\theta}$ does not depend on $(\sigma_f^2,\ell)$.
Recall
\begin{align*}
k_{\text{Matérn}}((s,a),(s',a')) = \sigma_f^2(1+\rho)\,e^{-\rho}, \quad
\rho = \frac{\sqrt{3}\,\|(s,a)-(s',a')\|}{\ell}.
\end{align*}
Let
\begin{align*}
\mathbf{K}_{\boldsymbol\phi} = [k_{\text{Matérn}}(x_i,x_j)]_{i,j}, \quad
\mathbf{K}_{\sigma} = \mathbf{K}_{\boldsymbol\phi} + \sigma_\epsilon^2\mathbf{I}, \quad
\mathbf{v} = \tilde{\mathbf{r}}-\boldsymbol\mu_{\boldsymbol\theta}, \quad
\boldsymbol\alpha = \mathbf{K}_{\sigma}^{-1}\mathbf{v}.
\end{align*}
Then for any hyperparameter $\phi \in \{\sigma_f^2, \ell\}$ we have the standard trace form:
\begin{align*}
\frac{\partial\mathcal{L}}{\partial\phi}
= \frac{1}{2}\operatorname{tr}\!\left(\left(\mathbf{K}_{\sigma}^{-1} - \boldsymbol\alpha\boldsymbol\alpha^\top\right)\frac{\partial\mathbf{K}_{\boldsymbol\phi}}{\partial\phi}\right).
\end{align*}

\paragraph{Gradient w.r.t. $\sigma_f^2$.}
Since
\begin{align*}
\frac{\partial k_{\text{Matérn}}}{\partial\sigma_f^2} = (1+\rho)e^{-\rho} = \frac{1}{\sigma_f^2}k_{\text{Matérn}},
\end{align*}
\begin{align*}
\frac{\partial\mathcal{L}}{\partial\sigma_f^2}
= \frac{1}{2\sigma_f^2}\left[\operatorname{tr}(\mathbf{K}_{\sigma}^{-1}\mathbf{K}_{\boldsymbol\phi})
- \boldsymbol\alpha^\top\mathbf{K}_{\boldsymbol\phi}\boldsymbol\alpha\right].
\end{align*}

\paragraph{Gradient w.r.t. lengthscale $\ell$.}
We have
\begin{align*}
\frac{\partial\rho}{\partial\ell} = -\frac{\rho}{\ell}, \qquad
\frac{d}{d\rho}\big[(1+\rho)e^{-\rho}\big] = -\rho e^{-\rho}.
\end{align*}
Hence, for each entry
\begin{align*}
\frac{\partial k_{ij}}{\partial\ell}
= \sigma_f^2\Big(-\rho e^{-\rho}\Big)\Big(-\frac{\rho}{\ell}\Big)
= \sigma_f^2\,\frac{\rho_{ij}^2}{\ell}\,e^{-\rho_{ij}}.
\end{align*}
Equivalently, in matrix/elementwise (Hadamard) form,
\begin{align*}
\frac{\partial\mathbf{K}_{\boldsymbol\phi}}{\partial\ell}
= \frac{1}{\ell}\,(\boldsymbol\rho\circ\boldsymbol\rho)\circ \mathbf{K}_{\boldsymbol\phi},
\end{align*}
where $[\boldsymbol\rho]_{ij}=\rho_{ij}$ and $\circ$ denotes Hadamard product. Therefore
\begin{align*}
\frac{\partial\mathcal{L}}{\partial\ell}
= \frac{1}{2}\operatorname{tr}\!\left(\left(\mathbf{K}_{\sigma}^{-1} - \boldsymbol\alpha\boldsymbol\alpha^\top\right)\frac{\partial\mathbf{K}_{\boldsymbol\phi}}{\partial\ell}\right).
\end{align*}
\noindent\emph{Note.} For $i=j$ we have $\rho_{ii}=0$, hence $[\partial\mathbf K_{\boldsymbol\phi}/\partial\ell]_{ii}=0$.

\subsection{Additional Kernel: Rational Quadratic Kernel Gradients}
We derive the gradients of the negative log-marginal-likelihood w.r.t. the Rational Quadratic kernel hyperparameters $\phi \in \{\sigma_f^2, \ell, \alpha\}$. Recall that
\begin{align*}
k_{\text{RQ}}(x,x') = \sigma_f^2\left(1 + \frac{\|x-x'\|^2}{2\alpha\ell^2}\right)^{-\alpha}, \quad x=(s,a), \; x'=(s',a').
\end{align*}
Define
\begin{align*}
\mathbf{K}_{\boldsymbol\phi} = [k_{\text{RQ}}(x_i,x_j)]_{i,j}, \quad \mathbf{K}_{\sigma} = \mathbf{K}_{\boldsymbol\phi} + \sigma_\epsilon^2\mathbf{I}, \quad \mathbf{v} = \tilde{\mathbf{r}}-\boldsymbol\mu_{\boldsymbol\theta}, \quad \boldsymbol\alpha = \mathbf{K}_{\sigma}^{-1}\mathbf{v}.
\end{align*}
Then for any kernel hyperparameter $\phi$,
\begin{align*}
\frac{\partial\mathcal{L}}{\partial\phi} = \frac{1}{2}\text{tr}\left(\left(\mathbf{K}_{\sigma}^{-1} - \boldsymbol\alpha\boldsymbol\alpha^\top\right)\frac{\partial\mathbf{K}_{\boldsymbol\phi}}{\partial\phi}\right).
\end{align*}

\paragraph{Gradient w.r.t. $\sigma_f^2$.}
\begin{align*}
\frac{\partial k_{ij}}{\partial\sigma_f^2} = \left(1 + \frac{d_{ij}^2}{2\alpha\ell^2}\right)^{-\alpha} = \frac{1}{\sigma_f^2}k_{ij},
\end{align*}
\begin{align*}
\frac{\partial\mathcal{L}}{\partial\sigma_f^2} = \frac{1}{2\sigma_f^2}\left[\text{tr}(\mathbf{K}_{\sigma}^{-1}\mathbf{K}_{\boldsymbol\phi}) - \boldsymbol\alpha^\top\mathbf{K}_{\boldsymbol\phi}\boldsymbol\alpha\right].
\end{align*}

\paragraph{Gradient w.r.t. length-scale $\ell$.}
Let $d_{ij} = \|x_i - x_j\|$ and $u=\frac{d_{ij}^2}{2\alpha\ell^2}$. Then
\begin{align*}
\frac{\partial}{\partial\ell}\left(1 + u\right)^{-\alpha}
&= -\alpha\left(1+u\right)^{-\alpha-1}\cdot \frac{\partial u}{\partial \ell},
\quad
\frac{\partial u}{\partial \ell}
= -\frac{d_{ij}^2}{\alpha \ell^3},
\\
\Rightarrow\quad
\frac{\partial}{\partial\ell}\left(1 + \frac{d_{ij}^2}{2\alpha\ell^2}\right)^{-\alpha}
&= \frac{d_{ij}^2}{\ell^3}\left(1 + \frac{d_{ij}^2}{2\alpha\ell^2}\right)^{-\alpha-1}.
\end{align*}
Hence
\begin{align*}
\frac{\partial k_{ij}}{\partial\ell}
= \sigma_f^2 \frac{d_{ij}^2}{\ell^3}
\left(1 + \frac{d_{ij}^2}{2\alpha\ell^2}\right)^{-\alpha-1}.
\end{align*}
Equivalently, since $k_{ij}=\sigma_f^2(1+u)^{-\alpha}$,
\begin{align*}
\frac{\partial k_{ij}}{\partial\ell}
= k_{ij}\,\frac{d_{ij}^2}{\ell^3}\,\frac{1}{1+ \frac{d_{ij}^2}{2\alpha\ell^2}}.
\end{align*}
Thus
\begin{align*}
\frac{\partial\mathcal{L}}{\partial\ell}
= \frac{1}{2}\operatorname{tr}\!\left(\left(\mathbf{K}_{\sigma}^{-1} - \boldsymbol\alpha\boldsymbol\alpha^\top\right)
\frac{\partial\mathbf{K}_{\boldsymbol\phi}}{\partial\ell}\right).
\end{align*}

\paragraph{Gradient w.r.t. shape parameter $\alpha$.}
\begin{align*}
\frac{\partial}{\partial\alpha}\left(1 + \frac{d_{ij}^2}{2\alpha\ell^2}\right)^{-\alpha} &= \left(1 + \frac{d_{ij}^2}{2\alpha\ell^2}\right)^{-\alpha}\left[-\ln\left(1 + \frac{d_{ij}^2}{2\alpha\ell^2}\right) + \frac{d_{ij}^2}{2\alpha\ell^2\left(1 + \frac{d_{ij}^2}{2\alpha\ell^2}\right)}\right].
\end{align*}
Simplifying:
\begin{align*}
\frac{\partial k_{ij}}{\partial\alpha} = k_{ij}\left[-\ln\left(1 + \frac{d_{ij}^2}{2\alpha\ell^2}\right) + \frac{d_{ij}^2}{2\alpha\ell^2 + d_{ij}^2}\right].
\end{align*}
Thus
\begin{align*}
\frac{\partial\mathcal{L}}{\partial\alpha} = \frac{1}{2}\text{tr}\left(\left(\mathbf{K}_{\sigma}^{-1} - \boldsymbol\alpha\boldsymbol\alpha^\top\right)\frac{\partial\mathbf{K}_{\boldsymbol\phi}}{\partial\alpha}\right).
\end{align*}

\end{document}